\documentclass[sigconf,natbib=true,anonymous=false]{acmart}
\AtBeginDocument{%
  }
\usepackage{multirow}
\setcopyright{cc}
\usepackage{xcolor}
\usepackage{colortbl}
\usepackage{booktabs}
\usepackage{multirow}
\usepackage{tcolorbox}
\usepackage{subfig}    
\usepackage{graphicx}  
\usepackage{float}     
\usepackage{epigraph}
\usepackage{algorithm}
\usepackage{algpseudocode}
\usepackage{fancyhdr} 
\newtcolorbox{cvbox}[1][]{  
    colback=gray!5,
    colframe=gray!75,
    boxrule=0.5pt,
    arc=2mm,
    left=5pt,
    right=5pt,
    top=5pt,
    bottom=5pt,
    title={#1},              
    fonttitle=\bfseries,     
    coltitle=white,          
    colbacktitle=gray!75,    
}

\definecolor{babygreen}{RGB}{220, 255, 220}
\definecolor{uclablue}{RGB}{220, 237, 255}
\definecolor{uclagold}{RGB}{255, 245, 220}
\definecolor{aliceblue}{RGB}{240, 248, 255}
\definecolor{thinkcolor}{RGB}{57,141,255}
\definecolor{toolcallcolor}{RGB}{255,127,14}
\definecolor{othercolor}{RGB}{153,153,153}
\definecolor{mygray}{RGB}{231, 231, 231}
\definecolor{increase}{RGB}{225, 182, 189}
\begin{document}
\fancyhead{}
\fancyhead[L]{DLLM-Searcher}
\fancyhead[RO]{GSAI IIR Lab.}         
\fancyhead[RE]{Zhao et al.}           

\title{DLLM-Searcher: Adapting Diffusion Large Language Models for Search Agents}

\author{Jiahao Zhao}
\authornote{Both authors contributed equally to this research.}
\author{Shaoxuan Xu}
\authornotemark[1]
\affiliation{%
  \institution{Renmin University of China}
  \city{Beijing}
  \country{China}
}
\email{zhaojiahao2202@ruc.edu.cn}

\author{Zhongxiang Sun}
\authornotemark[1]
\authornote{Project Leader}
\affiliation{%
  \institution{Renmin University of China}
  \city{Beijing}
  \country{China}
}
\email{sunzhongxiang@ruc.edu.cn}

\author{Fengqi Zhu}
\author{Jingyang Ou}
\affiliation{%
  \institution{Renmin University of China}
  \city{Beijing}
  \country{China}
}
\email{fengqizhu@ruc.edu.cn}

\author{Yuling Shi}
\affiliation{%
  \institution{Shanghai Jiao Tong University}
  \city{Shanghai}
  \country{China}
}
\email{yuling.shi@sjtu.edu.cn}

\author{Chongxuan Li}
\affiliation{%
  \institution{Renmin University of China}
  \city{Beijing}
  \country{China}
}
\email{chongxuanli@ruc.edu.cn}

\author{Jun Xu}
\authornote{Corresponding author.}
\author{Xiao Zhang}
\affiliation{%
  \institution{Renmin University of China}
  \city{Beijing}
  \country{China}
}
\email{junxu@ruc.edu.cn}
\renewcommand{\shortauthors}{Zhao et al.}

\begin{abstract}
Recently, Diffusion Large Language Models (dLLMs) have demonstrated unique efficiency advantages, enabled by their inherently parallel decoding mechanism and flexible generation paradigm. Meanwhile, despite the rapid advancement of Search Agents, their practical deployment is constrained by a fundamental limitation, termed as $\textbf{1) Latency Challenge}$: the serial execution of multi-round reasoning, tool calling, and tool response waiting under the ReAct agent paradigm induces severe end-to-end latency. Intuitively, dLLMs can leverage their distinctive strengths to 
optimize the operational efficiency of agents under the ReAct agent paradigm. Practically, existing dLLM backbones face the \textbf{2) Agent Ability Challenge}. That is, existing dLLMs exhibit remarkably weak reasoning and tool-calling capabilities, preventing these advantages from being effectively realized in practice. In this paper, we propose \textbf{DLLM-Searcher}, an optimization framework for dLLM-based Search Agents. To solve the Agent Ability Challenge, we design a two-stage post-training pipeline encompassing Agentic Supervised Fine-Tuning (Agentic SFT) and Agentic Variance-Reduced Preference Optimization $(\text{Agentic VRPO})$, which enhances the backbone dLLM's information seeking and reasoning capabilities. To mitigate the Latency Challenge, we leverage the flexible generation mechanism of dLLMs and propose a novel agent paradigm termed $\textbf{P}$arallel-$\textbf{Re}$asoning and $\textbf{Act}$ing $(\textbf{P-ReAct})$.
P-ReAct guides the model to prioritize decoding \texttt{tool\_call} instructions, thereby
allowing the model to \textit{keep thinking while waiting for the tool's return}. Experimental results demonstrate that DLLM-Searcher achieves performance comparable to mainstream LLM-based search agents
and P-ReAct delivers approximately 15\% inference acceleration. Our code is available at 
\textcolor{magenta}{\url{https://anonymous.4open.science/r/DLLM-Searcher-553C}}
\end{abstract}


\maketitle

\section{Introduction}

\begin{figure}[t] 
  \centering
       \includegraphics[width=1.0\linewidth]{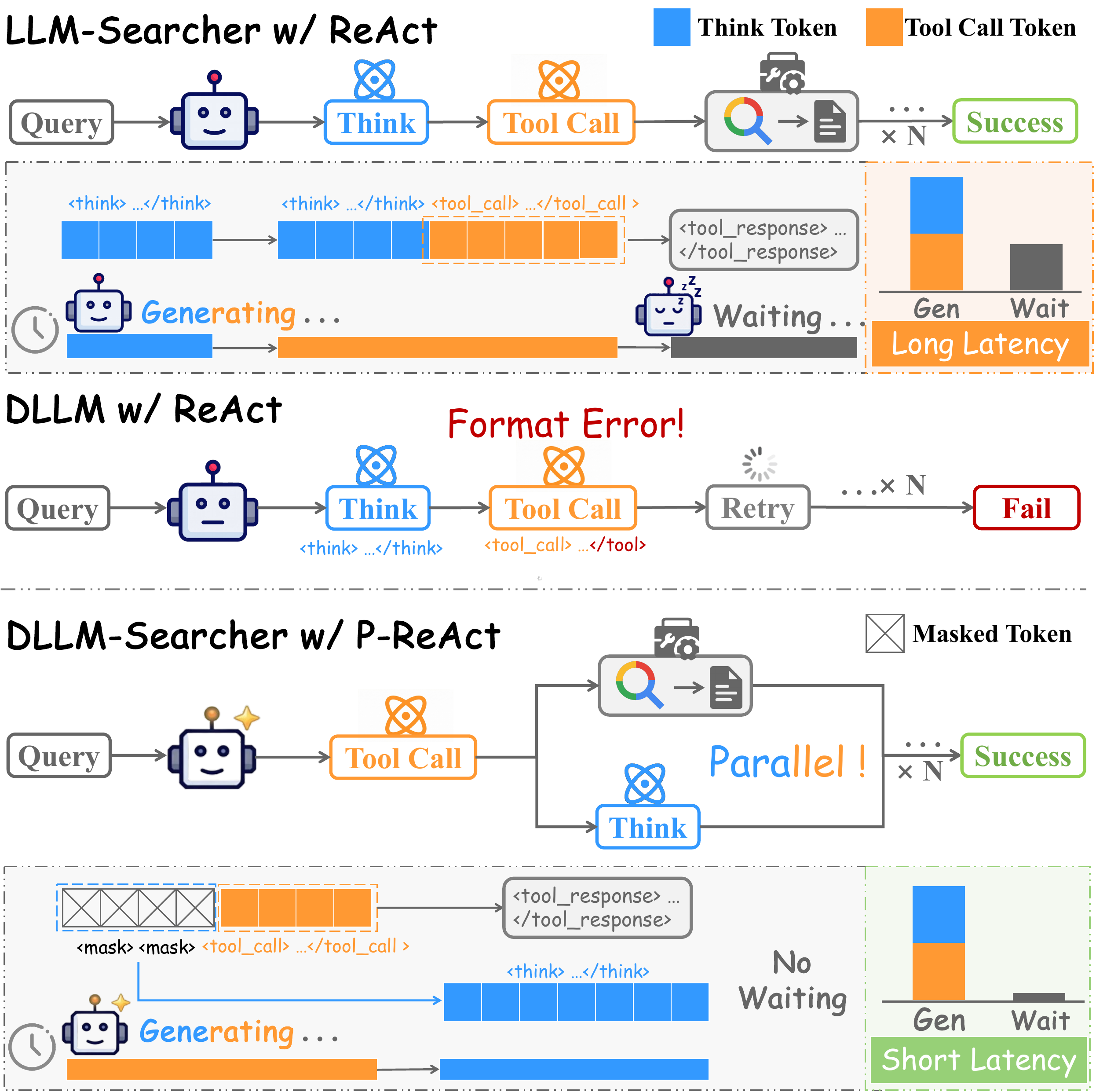} 
  \caption{LLM-based Search Agent (ReAct) vs. dLLM-based Search Agent (ReAct and P-ReAct). Top \& Middle: Standard ReAct paradigm suffers from high latency due to serial thinking and tool-calling regions generation and tool response waiting, while vanilla dLLMs fail due to tool-calling format errors. Bottom: P-ReAct prioritize tool-calling generation, enabling the model to keep thinking during tool execution.}
  \label{fig:Dllm-Searcher}    
  \vspace{-1.7em} 
\end{figure}
In recent years, Diffusion Large Language Models (dLLMs) have emerged as a promising alternative to traditional Autoregressive Models (ARMs)~\cite{LLaDA, dream7b, VRPO}. While ARMs are fundamentally constrained by a sequential ``left-to-right'' next-token prediction process, dLLMs leverage a non-causal diffusion mechanism that enables two
advantages: \textbf{parallel decoding mechanism} and \textbf{flexible generation paradigm}~\cite{LLaDA, dream7b, geminidiffusion, inception2025mercury, seeddiffusionlargescalediffusion, BDLM}. 

Concurrently, the integration of information retrieval with Large Language Models (LLMs) has led to the emergence of Search Agents, which enable LLMs to autonomously invoke tools to enhance generation quality~\cite{websailornavigatingsuperhumanreasoning, r1seacher, r1seacher++, smartsearcher}. 
Empowered by agentic post-training, these agents predominantly operate under the \textbf{Re}asoning and \textbf{Act}ing (\textbf{ReAct}) agent paradigm~\cite{yao2023reactsynergizingreasoningacting}. 
Under this paradigm, an agent first generates a \texttt{think} region to devise a search plan, followed by a \texttt{tool\_call} region to translate this plan into an API request and then halts generation to wait for external feedback. However, this serial execution creates a \textbf{1) Latency Challenge}: the end-to-end response time is severely bottlenecked by the cumulative delays of \texttt{think} and \texttt{tool\_call} generation and waiting for the tool response, as the model remains inactive during external tool execution.

Intuitively, dLLMs are ideal to mitigate this latency challenge. Beyond supporting parallel decoding for accelerated generation, their flexible generation paradigm offers significant potential for restructuring the ReAct execution flow. However, a significant gap exists between the theory and practice. As illustrated in \autoref{fig:Dllm-Searcher}, vanilla dLLMs frequently fail to adhere to specific tool-calling formats. Furthermore, their performance generally trails behind that of ARMs, particularly in agentic scenarios that demand robust reasoning and strict format compliance. These deficiencies constitute a critical \textbf{2) Agent Ability Challenge}, which hinders the practical deployment of dLLMs as search agent backbones.

In this paper, we propose \textbf{DLLM-Searcher}, an optimization framework that effectively enhances the information seeking and reasoning ability of dLLMs and improves their efficiency in agent scenarios. To address 2) Agent Ability Challenge, we design a two-stage post-training strategy, which is specifically tailored for improving dLLMs' agentic capability. To further exploit the strengths of dLLMs and boost the efficiency of dLLM-based search agents to deal with 1) Latency Challenge, we propose the \textbf{P}arallel-\textbf{Re}asoning and \textbf{Act}ing (P-ReAct), a novel agent paradigm that parallelizes the thinking and waiting phases.

\textbf{Training Process:} Specifically, our post-training pipeline is grounded in Agentic ELBO, a loss estimation method tailored for dLLM agents, and proceeds in two stages. First, we conduct Agentic \textbf{S}upervised \textbf{F}ine-\textbf{T}uning (\textbf{Agentic SFT}). This stage uses trajectories derived from a set of multi-hop questions to enable the model to acquire foundational information-seeking capabilities adapted to large block sizes. Subsequently, we employ Agentic \textbf{V}ariance-\textbf{R}educed \textbf{P}reference \textbf{O}ptimization (\textbf{Agentic VRPO}). By utilizing data filtered from post-SFT model rollouts, this stage further refines the model's reasoning and retrieval performance.

\textbf{Inference process:} The P-ReAct agent paradigm includes two critical components: Token Pre-filling and Confidence Biasing. We pre-fill \texttt{<tool\_call>} and \texttt{</tool\_call>} in the latter part of the block, and add a bias to the confidence scores of the positions between these two special tokens during the decoding process, guiding the model to prioritize decoding the content within this region.

We evaluate DLLM-Searcher along with existing Retrieval Augmented Generation (RAG) frameworks and search agents over four multi-hop benchmarks~\cite{hotpotqa, 2wiki, bamboogle, musique}. Experimental results show that the information seeking and reasoning capability of DLLM-Searcher is comparable to mainstream LLM-based search agents such as R1Searcher~\cite{r1seacher}. Moreover, it ensures with a nearly 100\% success rate that the tool-calling part is decoded first in every P-ReAct iteration process.

In summary, the contributions of our paper are as follows:
\begin{enumerate}
    \item \textbf{DLLMs Agent Post-training.} We develop a post-training pipeline for dLLMs, comprising Agentic SFT and Agentic VRPO, which can enhance dLLMs' agent capability.
    \item \textbf{DLLMs Agent Paradigm.} We propose \textbf{P-ReAct}, a novel, training-free paradigm adapted for dLLMs that guides the model to prioritize decoding high-quality tool calls, enabling parallel execution.
    \item \textbf{DLLM-Searcher Performance.} DLLM-Searcher achieves performance comparable to LLM-based search agents and realizes approximately 15\% acceleration compared to the ReAct paradigm.
\end{enumerate}
\section{Related Work}
\subsection{Diffusion Large Language Models}
\textit{Diffusion Language Models. } Inspired by discrete diffusion models~\cite{austin2021structured,campbell2022continuous,meng2023concrete,lou2023discrete,zhao2024improving}, dLLMs have emerged as a promising alternative to ARMs. LLaDA~\cite{LLaDA}, an 8B diffusion language model trained from scratch, achieves performance competitive with LLaMA3-8B. Dream7B~\cite{dream7b} introduces a comprehensive training framework that leverages AR-based LLM initialization and context-adaptive noise scheduling to scale diffusion language models. More recently, industrial efforts such as Gemini Diffusion~\cite{geminidiffusion}, Mercury~\cite{inception2025mercury}, and Seed-Diffusion~\cite{seeddiffusionlargescalediffusion} have further scaled dLLMs and demonstrated their potential for efficient inference.

\textit{Block Diffusion Language Models. } Furthermore, hybrid architectures such as Block Diffusion Language Models~(BDLMs)~\cite{BDLM, SDAR, Fastdllmv2, LLaDA2} have become a significant research focus. BDLMs employ an attention mechanisms combining intra-block bidirection with inter-block causal. This architecture natively supports KV Cache and variable-length text generation and keep the non-autoregressive generation ability within each block, allowing parallel decoding in arbitrary orders. SDAR~\cite{SDAR}, a series of BDLMs ranging from 1.7B to 8B parameters, has demonstrated general-purpose capabilities comparable to the latest open-source ARMs.

Despite these architectural advantages and competitive performance on tasks such as mathematics, existing dLLMs still lag behind ARMs in complex reasoning and agentic tasks. Weak reasoning capabilities and poor instruction-following ability prevent dLLMs from serving as agent backbone models. Our work enhances the agentic capabilities of dLLMs through a two-stage post-training pipeline, while leveraging their flexible generation mechanism to accelerate agentic inference.

\subsection{Search Agent}
Recent advances in search agents aim to integrate web search tool calling with the reasoning process of LLMs, enabling models to autonomously retrieve external knowledge after thinking. This paradigm significantly mitigates hallucination issues and enhances generation quality by grounding responses in retrieved evidence~\cite{shuster2021retrievalaugmentationreduceshallucination, gao2024retrievalaugmentedgenerationlargelanguage}. To further strengthen the synergy between reasoning and tool calling through training, researchers have explored diverse post-training strategies. R1-Searcher~\cite{r1seacher}, Search-R1~\cite{searchr1} employ a two-stage post-training pipeline consisting of Supervised Fine-Tuning (SFT) followed by Reinforcement Learning (RL) on open-source datasets, demonstrating substantial improvements in both reasoning and search capabilities. WebSailor~\cite{websailornavigatingsuperhumanreasoning} synthesize more challenging questions to push the boundaries of model search and reasoning abilities. MiroThinker~\cite{mirothinker} achieves superior performance by scaling the number of search iterations. \par
However, all these agents adopt the ReAct paradigm~\cite{yao2023reactsynergizingreasoningacting}, where reasoning, tool calling, and waiting for tool responses are executed serially. This sequential execution pattern forces users to endure prolonged waiting times, making latency a critical bottleneck for practical Search Agent deployment. DLLM-Searcher addresses this challenge by breaking the serial mechanism of ReAct, leveraging the flexible generation paradigm of dLLMs to enable parallel reasoning and action execution.

\section{Preliminary}
\subsection{Diffusion Large Language Models}
Formally, dLLMs model the data distribution through a forward-reverse framework. In the forward diffusion process, as the time step $t$ advances from 0 to 1, the clean input sequence $y$ is progressively corrupted by replacing tokens with a special mask token $[\texttt{M}]$ according to a transition probability $q_t$. Consequently, given the time step $t$ sampled uniformly from the interval $[0, 1]$, $y = (y^1, \dots, y^L)$ denote a clean input sequence of length $L$, and the conditioning prompt $x$, the transition probability $q_t$ is formulated as:
\begin{align}
  q_t(y_t \mid y, x) = \prod_{i=1}^{L} q_t(y_t^i \mid y^i, x),
  \\
  q_t(y_t^i \mid y^i, x) = 
  \begin{cases} 
    1 - t, & y_t^i = y^i, \\ 
    t, & y_t^i = [\texttt{M}]. 
  \end{cases}
  \label{eq:forward_process}
\end{align}

In the reverse process, the model predicts the original values of the masked tokens to compute the reverse probability $p_\theta(\cdot |y_t,x)$. To learn this distribution effectively, dLLMs adopts the Evidence Lower Bound (ELBO) $\mathcal{L}_\theta(y \mid x)$ as a surrogate objective to approximate the log conditional distribution $\log \pi_\theta(y|x)$~\cite{LLaDA, ESPO, RADD}:
\begin{align}
  \label{eq:elbo}
  \mathcal{L}_\theta(y \mid x) & \triangleq 
  \mathbb{E}_{t \sim \mathcal{U}[0,1] ,y_t \sim q_t(y_t \mid y, x)} \\ &\left[ 
    \frac{1}{t} \sum_{i=1}^{L} \mathbf{1}[y_t^i = [\texttt{M}]] \log p_\theta(y^i \mid y_t, x) 
  \right] \leq \log \pi_\theta(y \mid x). \nonumber
\end{align}
For BDLMs, the input $y$ is partitioned into $K$ continuous blocks  $[y^1, \dots,y^K]$, each of length $B$. The ELBO is defined as:
\begin{align}
    \label{eq:elbo_split}
    \mathcal{L}^{block}_\theta(y \mid x)  &\triangleq  \mathbb{E}_{t \sim \mathcal{U}[0,1], y_t \sim q_t(y_t \mid y, x)} \\
    &\Bigg[ \frac{1}{t} \sum_{k=1}^K \sum_{i=1}^{B} \mathbf{1}\Big[ y_t^{k,i} = [\texttt{M}] \Big] \log p_\theta(y^{k,i} \mid y_t^k,y^{<k}, x) \Bigg]. \nonumber
\end{align}


\subsection{Search Agents with ReAct}
In this section, we formally define the trajectory of a search agent interacting with the environment under the ReAct framework. Search Agents typically adopt ReAct as the agent framework. Let $f_{\theta}$ denote the agent LLM parameterized by $\theta$. Upon receiving a query $Q$ from the user, the agent follows the system prompt $S$ and performs several iterations of Thought-Action-Observation. 

We define the initial context as $\mathcal{H}_0 = (S, Q)$, which consists of the system prompt and the user query. At the $n$-th iteration ($n \geq 1$), let $T_n$, $A_n$, and $O_n$ denote the thought, action, and observation, respectively. Based on the existing context $\mathcal{H}_{n-1}$ from previous iterations, the agent generates a thought $T_n$ and executes a parsable action $A_n$:
\begin{equation*}
    (T_n, A_n) = f_{\theta}(\mathcal{H}_{n-1}),
\end{equation*}
then waits for the environment to return an observation $O_n$. The context is updated as $\mathcal{H}_n = (\mathcal{H}_{n-1}, T_n, A_n, O_n)$. In search scenarios, the action space consists of generating a final answer and calling the search tool with agent generated queries. The iteration terminates when the agent selects final answer as the action. 

Assuming the process terminates after $N$ iterations, the complete trajectory can be defined as:
\begin{equation*}
    \mathcal{H}_N = (S, Q, T_1, A_1, O_1, \ldots, T_i, A_i, O_i, \ldots, T_N, A_N).
\end{equation*}

\begin{figure*}[t]
  \centering
       \includegraphics[width=0.9\linewidth]{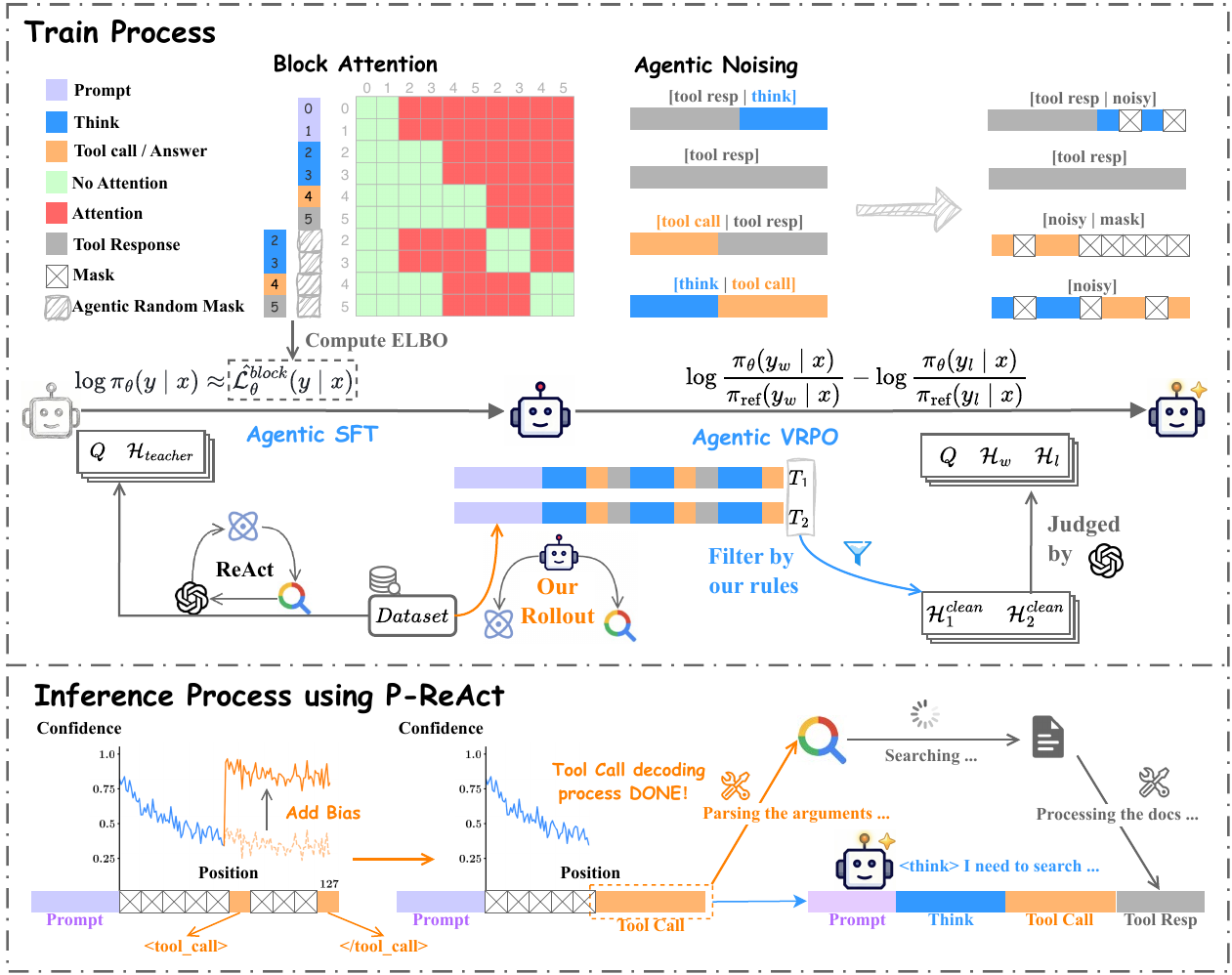} 
  \caption{DLLM-Searcher includes training process and inference process. In training, both Agentic SFT and Agentic VRPO use Block Attention and Agentic Noising to compute the Agentic ELBO, which serves to estimate $\log\pi_{\theta}(y \mid x)$. In inference, we employ the P-ReAct agent paradigm . We pre-fill special boundary tokens and apply an additional confidence bias to encourage the model to decode the \texttt{tool\_call} region with priority.}
  \label{fig:main}
\end{figure*}
\section{Our Approach: DLLM-Searcher} \label{section:method}
\subsection{Overview}
As illustrated in~\autoref{fig:main}, DLLM-Searcher consists of (i) a two-stage post-training pipeline and (ii) the P-ReAct agent paradigm.

\textit{Two-stage post-training pipeline.} Our experiments~\autoref{tab:main_results} show that dLLMs are weak in both multi-step reasoning and strict tool-call format following, which motivates Agentic post-training. \textbf{Agentic SFT~(\textsection~\ref{subsec:sft}).} We construct training data using trajectories generated by a stronger teacher model, in the form of $(Q, \mathcal{H}_{\text{teacher}})$. This stage improves the model's tool-call format following ability and helps it acquire initial capabilities to combine information retrieval with reasoning under large-block generation. \textbf{Agentic VRPO~(\textsection~\ref{subsec:vrpo}).} Starting from the SFT model, we roll out the trajectories and then filter them into winner/loser pairs $(Q, \mathcal{H}_w, \mathcal{H}_l)$ based on correctness. We then apply VRPO to further align the model toward correct trajectories, strengthening robust information-seeking behavior.

\textit{P-ReAct agent paradigm.} dLLMs can generate tokens in an arbitrary order within a block, but it's difficult to precisely control the generation order. Through extensive exploration, we develop a training-free strategy: we pre-fill the special boundary tokens \texttt{<tool\_call>} and \texttt{</tool\_call>} in the first step, and during subsequent decoding we apply a positional confidence bias to the token positions between these two boundaries. \textbf{P-ReAct~(\textsection~\ref{subsec:preact})} encourages the model to prioritize decoding the tool-call, effectively ensuring that tool-call instructions are generated ahead of the thinking process with near-perfect controllability.

\subsection{Agentic SFT}
\label{subsec:sft} 


We evaluate the agent capabilities of existing dLLM backbones and find that: dLLMs, particularly the BDLMs adopted in our work exhibit certain general-purpose abilities, they still fall short of the requirements in the Search Agent setting. Especially, they lack multi-step reasoning and tool-calling abilities. Therefore, we perform Agentic SFT to improve these capabilities. Furthermore, Search Agent trajectories typically contain external web content returned by search engines, whereas we want the model to learn only the \texttt{think} and \texttt{tool\_call} regions. Meanwhile, dLLMs are commonly optimized by maximizing the ELBO. To reconcile these characteristics, we propose \textbf{Agentic Noising} process and \textbf{Agentic ELBO} tailored for dLLM-based agent training.

\subsubsection{Data Construction}

Given a query $Q$, we use high-performance models to generate a teacher trajectory $\mathcal{H}_{\text{teacher}}$. We then apply a filter to retain only trajectories with correct final answers, clear and complete reasoning steps, and strictly valid tool call formats. The remaining query trajectory pairs $(Q, \mathcal{H}_{\text{teacher}})$ are used as training data.
\subsubsection{Block Attention and Agentic Nosing}

As illustrated in \autoref{fig:main}, the Block Attention used to train BDLMs adopts bidirectional attention within each block and causal attention across blocks. During training, we concatenate a noised trajectory after the clean one, forming an input of the form $[Q, \mathcal{H}_{\text{teacher}}, \mathcal{H}_{\text{noisy}}]$, which aims to perform calculations for all noisy blocks conditioned on their corresponding clean blocks in a single forward process under block attention. Since we want the model to learn only the \texttt{think} and \texttt{tool\_call} parts, we inject noise only into these components. Moreover, due to intra-block bidirectional attention, without additional intervention the model could access information from \texttt{tool\_response} tokens that appear later in the same block after the generation tokens, leading to a train inference mismatch. Therefore, in such cases we must fully mask the \texttt{tool\_response} tokens. We adopt the forward diffusion process $\hat{q}_t$ described above to noise $\mathcal{H}_{\text{teacher}}$ into $\mathcal{H}_{\text{noisy}}$, with the detailed procedure given in Algorithm~\autoref{alg:agentic_noising}.
\begin{algorithm}[t]
\caption{Agentic Noising Process $\hat{q}_t$}
\label{alg:agentic_noising}
\begin{algorithmic}[1]
\renewcommand{\algorithmicrequire}{\textbf{Input:}}
\renewcommand{\algorithmicensure}{\textbf{Result:}}
\Require sequence of tokens $y$, block size $B$, diffusion timestep $t$, set of tool response tokens $[\texttt{R}]$, mask token $[\texttt{M}]$
\Ensure noised sequence $y_t$ for training
\State Initialize $y_t \leftarrow y$
\State Partition $y$ into blocks $\{y^1, y^2, \dots, y^K\}$ of size $L$
\For{each block $y^k$}
    \State tool resp tokens' indices: $\mathcal{I}_{\text{resp}} \leftarrow \{i \in y^k \mid y^{(i)} \in [\texttt{R}]\}$
    \State gen tokens' indices: $\mathcal{I}_{\text{other}} \leftarrow \{j \in y^k \mid y^{(j)} \notin [\texttt{R}]\}$
    \If{$I_{\text{resp}} = \emptyset$}
        \State $\forall j \in y^k: y_t^{(j)} \sim q_t(y_t^{(j)} \mid y^{(j)}, x)$ \Comment{\textbf{1. Pure Gen Block}}
    \ElsIf{$I_{\text{other}} = \emptyset$}
        \State Skip \Comment{\textbf{2. Pure Response Block}}
    \Else
        \State $idx_R = \min(\mathcal{I}_{\text{resp}})$ 
        \State $idx_O = \min(\mathcal{I}_{\text{other}})$
        \If{$idx_R < idx_O$}
            \State  $\forall i \in \mathcal{I}_{\text{resp}}: y_t^{(i)} \leftarrow [\texttt{M}]$
            \State $\forall j \in \mathcal{I}_{\text{other}}: y_t^{(j)} \sim q_t(y_t^{(j)} \mid y^{(j)}, x)$ 
            \State \Comment{\textbf{3. Leakage Risk}}
        \Else
            \State $\forall i \in \mathcal{I}_{\text{resp}}: y_t^{(i)} \leftarrow y^{(i)}$ 
            \State $\forall j \in \mathcal{I}_{\text{other}}: y_t^{(j)} \sim q_t(y_t^{(j)} \mid y^{(j)}, x)$
            \State \Comment{\textbf{4. Observation Context}}
        \EndIf
    \EndIf
\EndFor
\State \Return $y_t$
\end{algorithmic}
\end{algorithm}
\subsubsection{Agentic ELBO}
Correspondingly, regarding the optimization objective, we aim for the model to focus on learning the \texttt{think} and \texttt{tool\_call} regions. Therefore, we adapt the Eq.~\ref{eq:elbo_split} to Agentic ELBO $\hat{\mathcal{L}}^{block}_\theta$ as follows:
\begin{align}
\label{eq:sft_loss_aligned} 
\hat{\mathcal{L}}&^{block}_\theta(y \mid x)  \triangleq  \ \mathbb{E}_{t \sim \mathcal{U}[0,1], y_t \sim \hat{q}_t(y_t \mid y, x)}  \\&
\Bigg[ \frac{1}{t} \sum_{k=1}^K\sum_{i=1}^{B} \mathbf{1}\Big[ (y_t^{k,i} = [\texttt{M}]) \land (y^{k,i} \notin [\texttt{R}]) \Big] \log p_\theta(y^{k,i} \mid y_t^k,y^{<k}, x) \Bigg], \nonumber
\end{align}
where we compute the loss only for tokens that are currently masked and were not originally in the \texttt{tool\_response} regions. Since our Agentic Noising may mask response tokens to prevent leakage, we exclude such positions from contributing to the loss.

\subsubsection{Training}
We set $y = \mathcal{H}_{\text{teacher}}$ and $x = S + Q$. In standard LLM training, the objective is typically the token-level negative log-likelihood $- \log \pi_\theta(y \mid x)$. In our setting, we use the proposed \textbf{Agentic ELBO} to approximate $\log \pi_{\theta}(y \mid x)$. Therefore, the final training loss is defined as the negative Agentic ELBO
\begin{align}
    \mathcal{L}_{SFT} =  \mathbb{E}_{(x, y) \sim \mathcal{D}} \big[- \hat{\mathcal{L}}^{block}_{\theta}(y \mid x)\big]. \nonumber
\end{align}

\subsection{Agentic VRPO}
\label{subsec:vrpo} 

Inspired by prior work on post-training dLLMs~\cite{VRPO}, VRPO can further improve model capability on top of SFT. Therefore, to further enhance the model’s reasoning and information retrieval abilities and to better adapt it to our P-ReAct agent paradigm, we introduce Agentic VRPO. Specifically, we roll out trajectories using the SFT model equipped with P-ReAct, and construct training data from these trajectories to train the model. During loss computation, we use the proposed Agentic ELBO introduced above to estimate $\pi_{\theta}(y \mid x)$ for both the reference model and the policy model.

\subsubsection{Data Construction}

Given a query $Q$, we perform two rollouts using the SFT model with P-ReAct to obtain two trajectories. We then select pairs where both trajectories are clear and complete and strictly follow the tool-call format, but one yields a correct final answer $\mathcal{H}_w$ while the other yields an incorrect one $\mathcal{H}_l$. The resulting training instance is formed as $(Q, \mathcal{H}_w, \mathcal{H}_l)$.

\subsubsection{Training}
Consistent with the SFT phase, we set $y_w = \mathcal{H}_w , y_l = \mathcal{H}_l, x = S + Q$, employ Agentic Noising $\hat{q}$ to ensure that tool responses do not disturbed the learning process. Consequently, we substitute the standard term in VRPO with our proposed Agentic ELBO $\hat{\mathcal{L}}^{block}_\theta(y \mid x)$. The final objective is formulated as:
\begin{align}
\mathcal{L}_{\text{VRPO}}(\theta) \triangleq \mathbb{E}_{(x, y_w, y_l) \sim \mathcal{D}} 
\Bigg[-\log \sigma \Big( \beta \Big[\Delta\mathcal{L}(y_w|x)- \Delta\mathcal{L}(y_l|x) \Big]\Big)\Bigg] , \nonumber
\end{align}
where $ \Delta\mathcal{L}(y|x) \triangleq \hat{\mathcal{L}}^{block}_\theta(y \mid x)-\hat{\mathcal{L}}^{block}_\text{ref}(y \mid x) $ represents the Agentic ELBO advantage of the policy over the reference model, $\beta$ is a hyperparameter that controls the deviation from the reference policy.

\subsection{P-ReAct Agent Paradigm}
\label{subsec:preact} 
The bidirectional attention mechanism in dLLMs allows them to access global context from tokens that have not yet been explicitly decoded~\cite{diffusionlanguagemodelsknow}. This provides a robust foundation for dLLMs to generate high-quality \texttt{tool\_call} instructions by leveraging information from the underlying reasoning trajectory even before the reasoning steps are fully decoded. Howerver, Our experiments reveal that without specific intervention, the generation order of dLLMs is stochastic and difficult to control. Notably, the latest BDLM backbone (SDAR) used in this work is finetuned from ARMs. It tends to degenerate into an autoregressive, left-to-right generation sequence within a block. To address this, we propose \textbf{P-ReAct}. We demonstrate that by pre-filling the two boundary special tokens for tool calls and applying a confidence bias to the span between them, we can prioritize the decoding of the \texttt{tool\_call} region with nearly 100\% probability. This enables the immediate parsing and dispatching of parameters to the search engine, while the model continues to generate the \texttt{think} component during the waiting period.


\subsubsection{Standard dLLMs Decoding}
We first formalize the standard inference process of dLLMs, utilizing the \textbf{Low-confidence Remasking} strategy. Given a prompt $x$, the model generates a response sequence $y$ of length $L$ over $N$ denoising steps. In general, $L=kN$, implying that $k$ new tokens are decoded at each step. Let $y_n$ denote the sequence state at step $n$, $\mathcal{M}_n$ be the set of masked indices at step $n$, and $\mathcal{V}$ be the vocabulary.

\noindent \textbf{Initialization}: Conventionally, the process begins with a fully masked sequence:
\begin{align*}
     y_0 = [\underbrace{\texttt{[M]}, \dots, \texttt{[M]}}_{L}], \quad \mathcal{M}_0 = \{1, \dots, L\}.
\end{align*}

\noindent \textbf{Denoising Step}: At each step $n$, the dLLM $f_\theta$ predicts logits $\mathbf{Z}_n = f_\theta(y_n, x) \in \mathbb{R}^{L \times |\mathcal{V}|}$. Only for a masked position $i \in \mathcal{M}_n$, we derive the probability distribution $P_\theta(y^i \mid y_n, x) = \text{Softmax}(z_n^i)$. We define the predicted token $\tilde{y}^i$ and its corresponding confidence score $C_n^i$ as follows:
\begin{align*}\tilde{y}^i = \mathop{\arg\max}\limits_{w \in \mathcal{V}} P_\theta(y^i = w \mid y_n, x), \
    C_n^i = \max_{w \in \mathcal{V}} P_\theta(y^i = w \mid y_n, x).
\end{align*}

The remasking strategy then selects a subset of positions with the highest confidence scores, unmasks them with their predicted tokens to form $y_{n+1}$, and updates $\mathcal{M}_{n+1}$.

\subsubsection{P-ReAct: Controlled Decoding Strategy}
The standard process described above implies an uncontrolled generation order. P-ReAct enforces a "Tool-First" hierarchy via two key modifications: \textbf{\texttt{tool\_call} Token Pre-filling} and \textbf{Confidence Biasing}.

\textbf{1. Special Token Pre-filling.}
To constrain the search space, we inject structural priors into the initialization $y_0$. Instead of a fully masked sequence, we pre-fill the boundary tokens for tool calls at designated positions. Let $pos_s$ and $pos_e$ denote the start and end indices for the tool span, respectively:

\begin{equation}
    \hat{y}_0^{(i)} = 
    \begin{cases}
        \texttt{<tool\_call>} & \text{if } i = pos_s, \\
        \texttt{</tool\_call>} & \text{if } i = pos_e, \\
        \texttt{[M]} & \text{otherwise}.
    \end{cases}
\end{equation}

By anchoring these boundaries, we explicitly define a structural skeleton in the noise space, forcing the model to generate valid tool content within the bracketed span.

\textbf{2. Confidence Biasing.}
To ensure the content enclosed by the anchors is decoded prior to the reasoning text, we adjust the confidence ranking step. 
Specifically, during the decoding iterations, we inject a positive bias $\alpha$ into the confidence scores of tokens located within the \texttt{tool\_call} region:
\begin{equation}
    \hat{C}_n^i = 
    \begin{cases}
    C_n^i + \alpha, & \text{if } pos_s < i < pos_e, \\
    C_n^i, & \text{otherwise}.
    \end{cases}
    \label{eq:confidence_bias}
\end{equation}
Given that the standard remasking strategy preferentially unmasks tokens with higher confidence, this bias effectively raises the decoding priority of the tokens within the \texttt{tool\_call} region, guaranteeing their generation in the earlier decoding steps.


\section{Experiments}
\begin{table*}[t]
\caption{Performance comparisons between Dllm-Searcher and the baselines on QA benchmarks. The best and second best results are \textbf{bold} and \underline{underlined}, respectively; ‘$^\dagger/\ddagger$’ represents in-domain/out-of-domain datasets; ‘$/$’ implies that the model struggles to generate valid tool call instructions, resulting in parsing failures; `$*$' means that the results were obtained under a modified experimental setup explained in \textsection~\ref{sec:baseline}).
}
\centering
\resizebox{0.85\linewidth}{!}
{
\begin{tabular}{lcccccccccc}
\toprule
\multirow{2}{*}{\textbf{Models}}  & \multicolumn{2}{c}{\textbf{HotpotQA$^\dagger$}} & \multicolumn{2}{c}{\textbf{2Wiki$^\dagger$}} & \multicolumn{2}{c}{\textbf{Bamboogle$^\ddagger$}} & \multicolumn{2}{c}{\textbf{Musique$^\dagger$}} & \multicolumn{2}{c}{\textbf{Avg}} \\
\cmidrule(lr){2-3} \cmidrule(lr){4-5} \cmidrule(lr){6-7} \cmidrule(lr){8-9} \cmidrule(lr){10-11}
  & \texttt{$ACC_R$} & \texttt{$ACC_L$} & \texttt{$ACC_R$} & \texttt{$ACC_L$} & \texttt{$ACC_R$} & \texttt{$ACC_L$} & \texttt{$ACC_R$} & \texttt{$ACC_L$} & \texttt{$ACC_R$} & \texttt{$ACC_L$} \\
\midrule
\multicolumn{11}{c}{\cellcolor{mygray} \textbf{\textit{Traditional RAG}}} \\
\midrule
SuRe & 32.4 & 48.4 & 22.2 & 26.8 & 17.6 & 28.0 & 7.2 & 10.0 & 19.9 & 28.3 \\
Selective-Context & 33.2 & 43.4 & 27.4 & 29.6 & 15.2 & 20.8 & 5.8 & 8.8 & 20.4 & 25.7 \\
Adaptive-RAG & 38.0 & 47.4 & 27.8 & 25.8 & 21.6 & 25.0 & 7.2 & 11.6 & 23.7 & 27.5 \\
IRCoT & 48.8 & 55.2 & 41.0 & 38.6 & 32.0 & 39.2 & 11.6 & 15.8 &  33.4& 37.2 \\
Iter-RetGen & 41.6 & 54.4 & 32.4 & 34.4 & 26.4 & 32.0 & 14.8 & 18.2 & 28.8 & 34.8 \\
CR-Planner & 44.4 & 33.6 & 48.2 & 22.0 & 35.2 & 34.4 & 12.2 & 11.4 & 35.0 & 25.4 \\
ReARTeR & 46.8 & 50.6 & 55.4 & 53.4 & 49.6 & 54.4 & \textbf{29.6} & \underline{30.2} & 45.4 & 47.2 \\
\midrule
\multicolumn{11}{c}{\cellcolor{mygray} \textbf{\textit{ARM-based LLMs Agent}}} \\
\midrule
Search-o1 & 40.8 & 53.2 & 47.0 & 51.2 & 49.6 & 52.0 & 15.2 & 19.0 & 38.2 & 43.9 \\
Search-R1 & 49.6 & \underline{62.2} & 46.0 & 50.0 & 47.2 & 56.0 & 28.0 & 26.0 & 42.7 & 48.6 \\
$\text{WebSailor}^*$ & 50.4 & 52.4  & 59.4 & 61.4 & 57.6 & 65.6  & 22.0 & 28.0 & 47.4 & 51.9  \\
$\text{R1Searcher}^*$ & \underline{58.0} & \underline{62.2} & \underline{59.6} & \underline{63.4} & \underline{66.4} & \underline{68.8} & 28.2  & \textbf{31.4} & \underline{53.1} & \underline{56.5} \\
\midrule
\multicolumn{11}{c}{\cellcolor{mygray} \textbf{\textit{dLLMs Agent}}} \\
\midrule
SDAR & / & / & / & / & / & / & / & / & / & / \\
Dream & 11.0 & 11.6 & 13.6 & 12.0 & 12.0 & 13.6 & 3.8 & 3.2 & 10.1 & 10.1\\
LLaDA & 36.0 & 32.8 & 42.0 & 38.8 & 46.4 & 42.4 & 15.2 & 15.8 & 34.9 & 32.5 \\
DLLM-Searcher & \textbf{60.4} & \textbf{62.4} & \textbf{69.8} & \textbf{64.6} & \textbf{68.8} & \textbf{69.6} & \underline{29.0} & 29.8 & \textbf{57.0} & \textbf{56.6} \\
\bottomrule
\end{tabular}
}
\label{tab:main_results}
\end{table*}


In this section, we empirically verify the effectiveness of DLLM-Searcher,
First, we conduct extensive comparisons between DLLM-Searcher and (i) traditional RAG methods, (ii) LLM-based agents, and (iii) dLLM-based agents, including our backbone SDAR, to verify that DLLM-Searcher improves the model’s information-seeking and reasoning capabilities. Then, to further analyze the effectiveness of the two core components of DLLM-Searcher, we formulate and answer the following research questions:

\textbf{RQ1: Effectiveness of the two-stage post-training pipeline.}
How does the proposed two-stage post-training pipeline, comprising Agentic SFT and Agentic VRPO, systematically enhance the information-seeking and reasoning abilities of dLLMs?

\textbf{RQ2: Inference efficiency brought by P-ReAct.}
How does P-ReAct achieve inference acceleration while maintaining performance?

\textbf{RQ3: The advantage of order-free generation.}
Is P-ReAct a capability unique to dLLMs? Can autoregressive LLMs generate the \texttt{tool\_call} region first without sacrificing performance?

Finally, we present a case study of a single P-ReAct iteration to qualitatively illustrate the ``thinking-while-waiting'' behavior exhibited by DLLM-Searcher in practice.

\subsection{Experimental Settings}
\subsubsection{Datasets}
This paper focuses on leveraging DLLM-Searcher to address complex multi-step question-answering (QA) tasks. To this end, four benchmark datasets are utilized in the experiments: HotpotQA~\cite{hotpotqa}, 2WikiMultiHopQA~\cite{2wiki}, Musique~\cite{musique}, Bamboogle~\cite{bamboogle}. Following the standard experimental setup of traditional RAG and
search agent~\cite{SKR, sure, flashrag, ReARTeR, r1seacher++, r1seacher,smartsearcher}, we sampled 500 examples from the development sets of HotpotQA, 2WikiMultiHopQA, and Musique as the test sets. For Bamboogle, which has only 125 examples in its test set and all of them are used in the experiments. 


To construct high-quality training data for DLLM-Searcher’s Agentic SFT, we design a trajectory sampling, rollout, and filtering pipeline. Specifically, we randomly sampling 2048 queries from each of the training sets of HotpotQA, 2WikiMultiHopQA, and Musique. Considering that Doubao-Seed-1.8 (251228)~\cite{doubao} is a recently released model with public API access, which demonstrates state-of-the-art performance in comprehensive capabilities, especially in search-related tasks, we utilize Doubao-Seed-1.8 to perform trajectory rollout with only one rollout iteration performed. Subsequently, we employ this model as the LLM judger with the prompt provided in our codebase. After that, we filter out trajectories that pass the LLM judge evaluation, feature complete reasoning paths, and comply with the standard \texttt{tool\_call} format, which are then used as training data for the Agentic SFT, resulting in a curated dataset of $3977$ trajectories. \par
For the Agentic VRPO, we utilize the SFT model to perform two rounds of rollouts on the 8k Stage 2 training samples released by R1Searcher. 
We then filter for queries where one rollout yields a correct answer while the other produces an incorrect one, with both corresponding trajectories being complete and format-compliant. This filtering process results in $2237$ qualified queries paired with $4474$ trajectories, which serve as the training data for the Agentic VRPO phase.



\subsubsection{Evaluation Metrics}
During evaluation, we observe that the outputs of search agents are typically long. Specifically, even when the model answers the question correctly, it often includes extensive supplementary information. As noted in prior work~\cite{r1seacher, ReARTeR}, this behavior makes exact-match metrics such as EM unsuitable for our setting. Following~\cite{r1seacher, ReARTeR}, we adopt accuracy ($\textbf{ACC}_{R}$) as our primary evaluation metric, which checks whether the golden answer is contained in the predicted answer generated by the search agent. To further refine our evaluation, we employ an LLM-as-Judge protocol~\cite{llmsasjudge} using Doubao-seed-1.8 as the judge model to determine whether the predicted answer is correct, denoted as $\textbf{ACC}_{L}$.

\subsubsection{Baselines}
\label{sec:baseline}
To verify the effectiveness of DLLM-Searcher in enhancing the reasoning and information seeking capabilities of dLLMs, We compared DLLM-Searcher against several baselines:  \\
\textbf{Traditional RAG :}
\textit{SuRe}~\cite{sure} executes multiple reasoning paths in parallel for a single query. \textit{Selective-Context}~\cite{selective} compresses retrieved documents to reduce context length. \textit{Adaptive-RAG}~\cite{SKR} dynamically selects retrieval strategies depending on the complexity of the query. \textit{RAG-CoT methods}, such as IRCoT~\cite{IRCOT},  Iter-RetGen~\cite{Iterretgen}. \textit{CR-Planner}~\cite{CRPlanner}, ReARTeR~\cite{ReARTeR} scales RAG at inference time using Monte Carlo Tree Search (MCTS).\\
\textbf{LLM Agents :} \textit{Search-o1}~\cite{searcho1} integrates RAG with Chain-of-Thought (CoT) reasoning via prompt engineering. 
For models that leverage reinforcement learning (RL) to autonomously learn retrieval behaviors during inference, we include \textit{Search-r1}~\cite{searchr1}, \textit{WebSailor}~\cite{websailornavigatingsuperhumanreasoning}, and \textit{R1Searcher}~\cite{r1seacher} as baselines. 
Note that WebSailor was trained with two tools, namely \textit{search} and \textit{visit}. To ensure consistency across all evaluations, we only equip it with the \textit{search} tool in our experiments. 
R1Searcher was trained using a local search tool, the results reported correspond to the higher performance achieved between evaluations with the local search and Google Search. \\
\textbf{dLLM Agents :} To quantitatively benchmark the intrinsic performance of dLLMs in agentic tasks, we directly evaluate the dLLM backbone \textit{SDAR}~\cite{SDAR}, \textit{Dream}~\cite{dream7b}, \textit{LLaDA}~\cite{LLaDA} using the standard ReAct paradigm, what's more, for LLaDA and Dream, we use Fast-dLLM~\cite{wu2025fastdllmtrainingfreeaccelerationdiffusion} to accelerate the inference.

\subsubsection{Implementation Details} 
\noindent\textbf{Model and Tools.} We employ the SDAR model with a block size of 64 as our backbone. For the retrieval component, we utilize Google Search as our external tool, retrieving the top 10 search results. 

\noindent\textbf{Agentic SFT.} During the SFT stage, we utilize an attention mask with a block size of 128. The training is conducted with a learning rate of $1e^{-5}$, a total batch size of $32$, and for $3$ epochs. 

\noindent\textbf{Agentic VRPO.} In the VRPO stage, we maintain an attention mask block size of 128. The model is trained with a learning rate of $5e^{-7}$ and a batch size of $16$ for $5$ epochs.

\noindent\textbf{Decoding Configuration.} For both the VRPO rollout data generation and final evaluation, we apply our proposed P-ReAct strategy. We set the \textit{confidence bias} to 0.5 and employ a \textit{low-confidence static} approach for the remaining decoding steps. The inference configuration consists of 128 denoising steps, a block size of 128, and a temperature of $1.0$.

\noindent\textbf{Hardware.} All experiments and training processes are implemented using the PyTorch framework and conducted on a server equipped with 8 $\times$ NVIDIA H100 GPUs.
\subsection{Overall Performance}
Table\ref{tab:main_results} shows the results of DLLM-Searcher and the baselines on four mutil-hop QA benchmarks. We can obtain the following observations: 

\emph{\textbf{1. DLLM-Searcher achieved performance improvements on multi-hop QA.}}
Our method, DLLM-Searcher, achieves excellent performance across all multi-hop QA benchmarks under both the $\boldsymbol{ACC_R}$ and $\boldsymbol{ACC_L}$ metrics. \par
(1) It outperforms traditional RAG strategies by a substantial margin, especially attaining an improvement of about 19\% over ReARTeR which is a strong baseline that leverages a PRM model to supervise the reasoning process.(2) It yields significant performance gains compared with vanilla dLLMs without any agentic post-training.(3) It achieves comparable performance against search agents built on ARMs, with the only performance gap observed on the Musique dataset relative to R1Searcher. \par
These results demonstrate that our two-stage post-training strategy tailored for dLLMs effectively enables the model to perform accurate and timely retrieval invocations throughout the reasoning process, thereby enhancing overall performance.


\emph{\textbf{2. Maintaining Generalization Ability.}} 
Despite being trained on only 8k samples, DLLM-Searcher achieves strong performance on in-domain datasets such as HotpotQA, 2WikiMultiHopQA, and Musique, while also demonstrating impressive generalization capability on out-of-domain datasets such as bamboogle. This indicates that the model has effectively learned to integrate the retrieval of relevant documents with its internal reasoning process during training. Such an ability ensures the model's robust performance on unseen datasets that require external information retrieval.

Furthermore, all results of DLLM-Searcher presented in Table~\ref{tab:main_results} are obtained using the P-ReAct paradigm, which achieves substantial inference efficiency gains over the standard ReAct paradigm with negligible performance degradation. More results are reported in Section \textsection~\ref{sec:rq2}.

\subsection{Empirical Analysis}
We conducted experiments to analyze the components of DLLM-Searcher and answer the following research questions. 

\subsubsection{\textbf{RQ1: Effectiveness of Post-training.}} 

\begin{table}[t]
\centering
\caption{Performance comparison between Agentic SFT and Agentic VRPO on different datasets}
\label{tab:sft_dpo}
\begin{tabular}{llcr@{\hspace{2pt}}l}
\toprule
\multirow{2}{*}{\textbf{Dataset}} & \multirow{2}{*}{\textbf{Metric}} & \multicolumn{3}{c}{\textbf{Method}} \\
\cmidrule(lr){3-5}
 &  & \textbf{\quad \quad SFT \quad \quad} & \multicolumn{2}{c}{\textbf{VRPO \quad \quad}} \\
\midrule
\multirow{2}{*}{\textbf{HotpotQA}} & \texttt{$ACC_R$} & 57.2 & 60.&4 $_{{(+3.2)}}$  \\
 & \texttt{$ACC_L$} & 58.8 & 62.&4 $_{{(+3.6)}}$  \\
\midrule
\multirow{2}{*}{\textbf{2Wiki}} & \texttt{$ACC_R$} & 66.4 & 69.&8 $_{{(+3.4)}}$  \\
 & \texttt{$ACC_L$} & 61.6 & 64.&6 $_{{(+3.0)}}$ \\
\midrule
\multirow{2}{*}{\textbf{Bamboogle}} & \texttt{$ACC_R$} & 64.6 & 68.&8 $_{{(+4.2)}}$  \\
 & \texttt{$ACC_L$} & 64.0 & 69.&6 $_{{(+5.6)}}$  \\
\midrule
\multirow{2}{*}{\textbf{Musique}} & \texttt{$ACC_R$} & 24.4 & 29.&0 $_{{(+4.6)}}$ \\
 & \texttt{$ACC_L$} & 26.6 & 29.&8 $_{{(+3.2)}}$ \\
\bottomrule
\end{tabular}
\vspace{-1em}
\end{table}
We first evaluate the instruction-following capabilities of the vanilla SDAR model in agentic scenarios. We find that the model exhibits an almost complete inability to adhere to the rigid formatting protocols required for agentic interaction. 

Specifically, when tested on 500 samples from the HotpotQA dataset, the vanilla SDAR model failed to complete a single successful interactions. \textbf{All test cases} were terminated during the first turn of the ReAct process due to formatt errors. We have summarized the 4 most frequently occurring types of errors:
\textbf{1. Empty Output}: The model directly outputs the end token "\texttt{<|im\_end|>}" without generating any response content.
\textbf{2. No \texttt{tool\_call}}: The model generates reasoning process within the \texttt{<think>} tag but fails to produce the required \texttt{tool\_call} afterwards.
\textbf{3. \texttt{think} Format Error}: The model produces incomplete or malformed \texttt{think} tags, such as unclosed tags like \texttt{<th}.
\textbf{4. \texttt{tool\_call} Format Error}: The model generates \texttt{tool\_call} with incorrect JSON syntax or malformed function call structure like \texttt{<tools>}.
The detailed statistics are presented in Table~\ref{tab:error_classification}. 

We attribute this catastrophic failure to the absence of instruction data tailored for tool-use and multi-step reasoning during the SDAR's Continue Pre-Train (CPT) phase. Without targeted alignment specific to agentic workflows, the model suffers from severe structural hallucinations. 

\begin{table}[t]
\centering
\caption{Error Type Classification and Statistics}
\label{tab:error_classification}
\begin{tabular}{lcc}
\toprule
\textbf{Error Type} & \textbf{Count} & \textbf{Percentage} \\
\midrule
\textbf{Empty Output} & 156 & 31.20\% \\
\midrule
\textbf{No \texttt{tool\_call}} & 142 & 28.40\% \\
\midrule
\textbf{\texttt{think} Format Error} & 89 & 17.80\% \\
\midrule
\textbf{\texttt{tool\_call} Format Error} & 35 & 7.00\% \\
\bottomrule
\end{tabular}
\vspace{-1em}
\end{table}

As reported in Table~\ref{tab:sft_dpo}, Agentic SFT effectively rectifies these behavioral flaws, enabling SDAR to strictly follow the requisite \texttt{tool\_call} formats. Consequently, the reasoning trajectories are no longer prematurely terminated early by parsing failures, allowing the model to successfully execute multi-step logic chains. The quantitative results across all four datasets confirm that the post-SFT model has acquired fundamental information seeking and reasoning capabilities. Furthermore, the subsequent Agentic VRPO stage delivers additional performance gains, consistently enhancing the model's capabilities and yielding universal improvements across all benchmarks. Remarkably, both $ACC_R$ and $ACC_L$ exhibit gains exceeding 3 percentage points.

\subsubsection{\textbf{RQ2: Inference Efficiency.}}
\label{sec:rq2}
\begin{figure}[t] 
  \centering
  \includegraphics[width=1.0\linewidth]{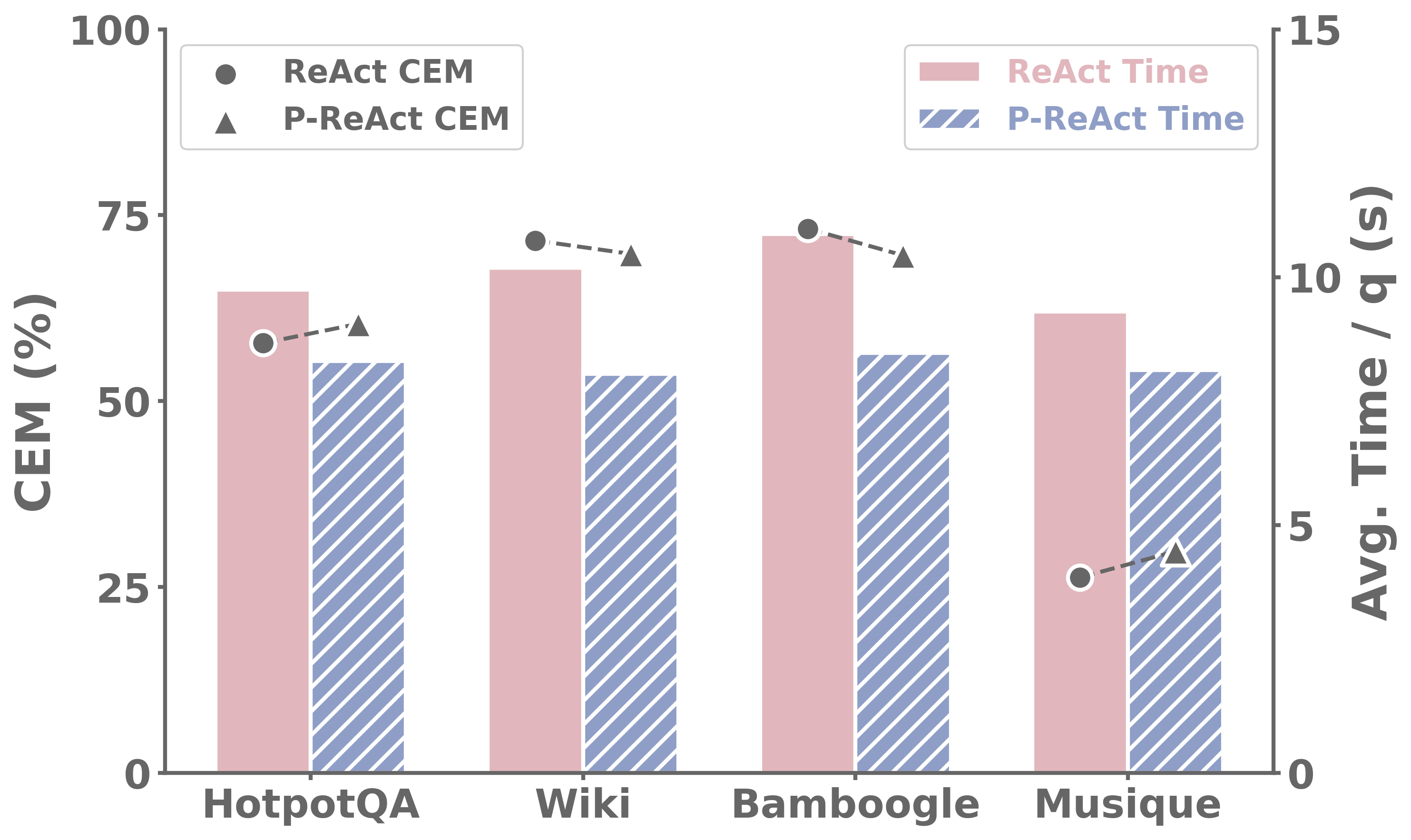} 
  \caption{Accuracy and average end-to-end latency comparison of DLLM-Searcher under P-ReAct and ReAct paradigms on multi-hop QA tasks.}
  \label{fig:P-ReAct vs. ReAct}
  \vspace{-0.9em}
\end{figure}
We evaluate our final model using both ReAct and P-ReAct. Under ReAct, we impose no additional constraints and allow the model to follow the standard Reasoning-Action-Observation cycle until reaching the maximum number of turns or producing a final answer. Under P-ReAct, we constrain each turn to complete the \texttt{think} and \texttt{tool\_call} region within a single block, by pre-filling the \texttt{<tool\_call>} boundary tokens and applying a confidence bias to guide decoding. As shown in the~\autoref{fig:P-ReAct vs. ReAct}, across the four datasets, P-ReAct achieves inference time reductions of 14.77\%, 21.00\%, 22.08\%, and 12.67\% relative to ReAct, with almost no performance degradation. These results indicate that P-ReAct effectively exploits the properties of dLLMs to prioritize decoding high-quality tool calls, and further accelerates search agent inference by overlapping reasoning with the waiting time for tool responses.

\subsubsection{\textbf{RQ3: Advantages of dLLMs' Order-free Generation.}}
\begin{figure}[t] 
  \centering
  \includegraphics[width=1.0\linewidth]{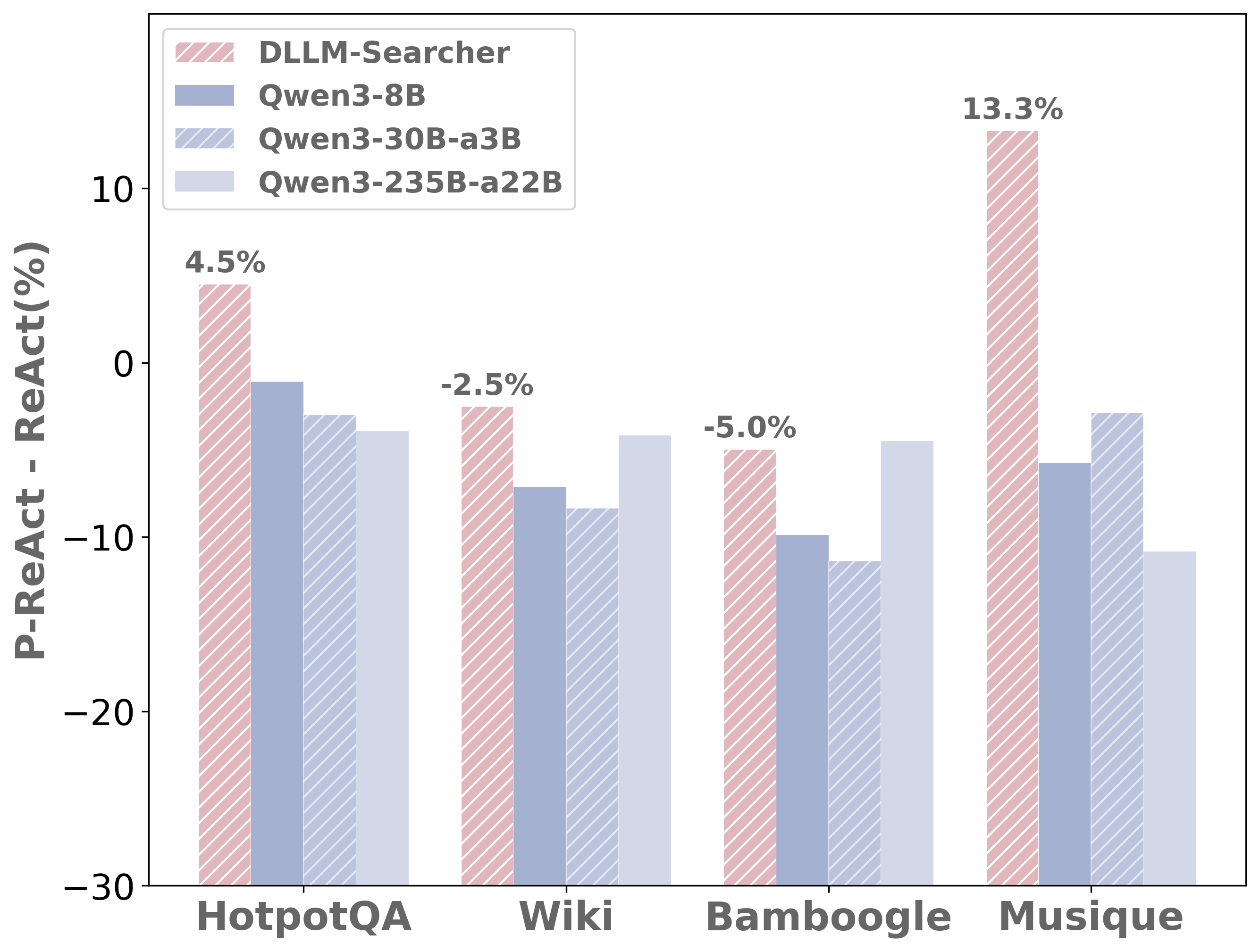} 
  \caption{Comparison of accuracy changes on multi-hop QA tasks between DLLM-Searcher and Qwen3 series models when switching from ReAct to P-ReAct.}
  \label{fig:Reverse ReAct}
  \vspace{-1em}
\end{figure}

Constrained by the causal attention mechanism and the next-token prediction paradigm, LLMs can only achieve parallel ``thinking while waiting for tool response'' capabilities comparable to P-ReAct in a training-free setting by restructuring each ReAct output into the sequence [\texttt{<tool\_call>}...\texttt{</tool\_call>} \texttt{<think>}...\texttt{</think>}]. We conducted experiments using three models of varying sizes from the Qwen3 series : Qwen3-8B, Qwen3-30B-A3B, and Qwen3-235B-A22B, by modifying the prompts to instruct the models to generate the \texttt{tool\_call} component first. As shown in \autoref{fig:Reverse ReAct}, ARMs can only achieve this behavior at the cost of notable performance degradation. In contrast, DLLM-Searcher with P-ReAct even achieves accuracy gains over standard ReAct on the HotpotQA and Musique datasets, with only minor accuracy losses observed on 2Wiki and Bamboogle. Overall, its performance degradation is far less significant than that of ARMs.

This experiment further demonstrates that ARMs rely heavily on explicitly decoded \texttt{think} segments to generate high-quality \texttt{tool\_call} instructions. In contrast, although DLLM-Searcher with P-ReAct ostensibly decodes the \texttt{tool\_call} component first, the quality of the generated tool calls remains well-preserved. As discussed in \textsection~\ref{subsec:preact}, diffusion models ``know the answer before decoding it'' they leverage bidirectional attention within blocks such that, even when the \texttt{think} region is not explicitly decoded, the model still utilizes the relevant reasoning information during the \texttt{tool\_call} decoding process. In essence, each decoding step implements a form of latent reasoning, enabling robust tool call generation without explicit chain-of-thought verbalization.

\subsection{Case Study}
In this section, we present a real case to show how DLLM-Searcher enhances the performances. Specifically, we showed the first thinking and tool calling iteration of both the vanilla model and DLLM-Searcher on a HotpotQA query: \texttt{"In the 1973 NFL season, the Pro Bowl took place at what football stadium that is part of the Truman Sports Complex, along with what other stadium?"}. As observed, the vanilla model outputs an incorrect special token \texttt{<tool>}, whereas DLLM-Searcher produces the correct format. Furthermore, at timestep 32, only the \texttt{tool\_call} region is fully decoded, while the \texttt{think} region remains entirely masked. The \texttt{think} region is not fully populated until the final stage of decoding. These observations demonstrate that our two-stage post-training procedure, combined with the P-ReAct agent paradigm, significantly enhances the model's instruction-following capability. Additionally, both reasoning and information-seeking abilities are notably improved, and the model successfully prioritizes the decoding of the \texttt{tool\_call} component. This further validates the effectiveness of our proposed P-ReAct strategy.
\begin{cvbox}[Vanilla Model's One ReAct Iteration]
\small\ttfamily
First, let's identify the Pro Bowl's location for 1973 and also verify the stadium, so we can compare it with the Truman Sports Complex. We'll perform our search for these details using our search tool.\textbackslash{}n\textbackslash{}n
\textbf{\color{red}{<tool>}}\textbackslash{}n
\{"name": "search", "arguments": \{"query": ["Pro Bowl 1973 location", "Truman Sports Complex arenas"]\}\}\textbackslash{}n
\textbf{\color{red}{</tool>}}\textbackslash{}n
<|im\_end|>\textbackslash{}n<|endoftext|>
\end{cvbox}

\begin{cvbox}[DLLM-Searcher's One P-ReAct Iteration]
\small\ttfamily
<think> \textbackslash{}n To answer the question, I need to find the location of the 1973 NFL Pro Bowl and confirm if it's within the Truman Sports Complex. Then, I need to identify the other stadium in that complex. First, I'll search for the 1973 NFL's Pro Bowl location and check if it's in the Truman Sports Complex. If so, I'll search for the other stadium in that complex to provide the full answer. \textbackslash{}n </think>\textbackslash{}n \textbackslash{}n
\textcolor{toolcallcolor}{<tool\_call> \textbackslash{}n \{ "name": "search", "arguments": \{ "query": [ "1973 NFL Pro Bowl location", "Truman Sports Complex Pro Bowl stadiums list" ] \} \textbackslash{}n \} </tool\_call>}
<|im\_end|>
\end{cvbox}

\begin{cvbox}[Decoding Order]
\small\ttfamily
35 33 37 39 38 42 41 40 43 44 45 55 58 57 46 47 48 49 51 50 53 54 52 56 59 64 65 66 67 63 62 61 60 68 74 69 70 72 71 73 78 75 76 77 88 79 80 81 82 83 87 84 85 86 95 91 92 94 93 96 97 90 89 98 99 109 100 102 101 108 107 105 104 103 106 112 123 110 111 113 114 115 116 117 121 118 125 120 119 126
\textcolor{toolcallcolor}{1 3 2 5 4 7 6 17 8 16 9 10 12 13 15 14 18 20 19 21 26 24 25 22 23 28 29 30 31 11 3 1 27}
124
\end{cvbox}

\section{Conclusion}
This paper presents \textbf{DLLM-Searcher} to bridge the gap between Diffusion Large Language Models and practical Search Agents. Two major obstacles that prevent this adoption are analyzed: the \textit{Agent Ability Challenge} stemming from the dLLM backbone, and the \textit{Latency Challenge} arising from the conventional ReAct paradigm. To address these challenges, we propose a tailored two-stage post-training pipeline comprising \textbf{Agentic SFT} and \textbf{Agentic VRPO}, along with a novel \textbf{P-ReAct} paradigm. In this way, DLLM-Searcher enables dLLMs to \textit{keep thinking while waiting} during external tool execution. Experimental results on four benchmarks demonstrate that DLLM-Searcher achieves approximately 15\% inference acceleration over the conventional ReAct paradigm while maintaining comparable performance to mainstream ARM-based search agents, verifying the potential of dLLMs as efficient agent backbones and the effectiveness of parallelizing agentic reasoning and acting.
\begin{flushright}
\textit{``We actually start to act before we are aware of our decision to do so.''} \\
\vspace{0.3em}
--- Ray Kurzweil, \textit{How to Create a Mind}
\end{flushright}

\section{Acknowledgements}
We thank the group of Dr. Biqing Qi at Shanghai AI Lab for their work on SDAR: an excellent BDLM. We also appreciate the guidance from Dr. Biqing Qi and Shuang Chen during the SFT stage: modifying the block attention to use a block size of 128 and then training normally. We also thank Jinhao Jiang, Huatong Song and Peitian Zhang for their valuable insights on the training of search agents.
\bibliography{sample-base}

\end{document}